\documentclass[3p,times,procedia]{elsarticle}
\usepackage{graphicx}
\usepackage{amsmath}
\usepackage{xcolor}
\usepackage{booktabs}
\usepackage{caption}
\usepackage{textalpha} 
\usepackage{subcaption}

\usepackage[bookmarks=false]{hyperref}
    \hypersetup{colorlinks,
      linkcolor=blue,
      citecolor=blue,
      urlcolor=blue}

\usepackage{amssymb}

\usepackage[figuresright]{rotating}

\begin{document}
\begin{frontmatter}





\title {Intelligent System for Assessing University Student Personality Development and Career Readiness}


\author[a]{Izbassar Assylzhan} 
\author[a]{Muragul Muratbekova}
\author[a]{Daniyar Amangeldi}
\author[a]{Nazzere Oryngozha} 
\author[a]{Anna Ogorodova} 
\author[a]{Pakizar Shamoi\corref{cor1}}

\address[a]{School of Information Technology and Engineering, Kazakh-British Technical University, Almaty, Kazakhstan}

\begin{abstract}
While academic metrics such as transcripts and GPA are commonly used to evaluate students' knowledge acquisition, there is a lack of comprehensive metrics to measure their preparedness for the challenges of post-graduation life. This research paper explores the impact of various factors on university students' readiness for change and transition, with a focus on their preparedness for careers. The methodology employed in this study involves designing a survey based on Paul J. Mayer's "The Balance Wheel" to capture students' sentiments on various life aspects, including satisfaction with the educational process and expectations of salary. The collected data from a KBTU student survey (n=47) were processed through machine learning models: Linear Regression, Support Vector Regression (SVR), Random Forest Regression. Subsequently, an intelligent system was built using these models and fuzzy sets. The system is capable of evaluating graduates' readiness for their future careers and demonstrates a high predictive power. The findings of this research have practical implications for educational institutions. Such an intelligent system can serve as a valuable tool for universities to assess and enhance students' preparedness for post-graduation challenges. By recognizing the factors contributing to students' readiness for change, universities can refine curricula and processes to better prepare students for their career journeys.

\end{abstract}

\begin{keyword}
assessment system, linear regression, random forest regression, support vector regression, fuzzy sets, machine learning.




\end{keyword}
\cortext[cor1]{Corresponding author. Tel.: +7-701-349-0001.}
\end{frontmatter}




\section{Introduction} 



Many universities are highly focused on academic assessment \cite{perf}. 
Although academic measures like transcripts and GPA are frequently used to assess students'  knowledge, there is a lack of metrics for evaluating critical qualities like readiness for the career and challenges of post-graduation life \cite{org}. 

The focus of the present study is to shed light on this underexplored dimension of university education. We explore the impact of various factors on university students' readiness for change, with a particular focus on their preparedness for career development. Our aim is to integrate the ML-based fuzzy intelligent system into the university platform so that examining students' openness to new opportunities and their perceptions of career readiness can be better tracked, and intervention can be made timely.

Educational institutions worldwide provide students with guidance and counseling services to help them effectively navigate life's challenges and obstacles. Researchers have extensively analyzed the factors that may affect students' academic performance \cite{new1, covid19}. Previous studies have used correlation analysis, decision trees, and random forests to identify factors that affect student performance \cite{success}.

Fuzzy sets were used to characterize students' state of knowledge, highlighting the lack of collaborative  support and feedback in e-learning systems \cite{profiling}. Similarly, other researchers have compared recommender systems methods with traditional regression methods, applying them to educational data for intelligent tutoring systems to improve the accuracy of predicting student performance \cite{perf}. The authors \cite{fuzzycool} discussed building a semantic student profile and using it to predict students' learning preferences based on their interests and learning style using fuzzy logic and the results of an online questionnaire that uses only knowledge about students' interests and style. Next, some authors emphasized the applicability of fuzzy sets to data representation, arguing that self-awareness is crucial to their model \cite{dubai, fuzzycool}. 
Several studies have proposed systems that identify different stages of dropout and offer motivational phrases and advice to support students in different scenarios \cite{retention}. The system, based on comprehensive information about students' grades in different disciplines, produces results to improve the efficiency and responsiveness of decision-making \cite{case} to improve the education system at the university.

The results of the survey among students showed a positive attitude towards the implementation of electronic systems in student life \cite{ready}. Researches on mental health have found that it is possible to use ML to identify subjective well-being and potential risks of mental health problems of students \cite{mental, turks}. ML has also been used to understand the readiness of students from different groups to promote information education \cite{python}. Similarly, the author used multiple regression analysis to identify the correlation between emotional intelligence and organization change readiness \cite{org}. The intelligent system to identify student progress was developed using the various methodologies such as Big Five Factors and Five Factor Model, intelligence quotient tests, self-assessment \cite{bio}. 

 The contributions of this work include dataset collection, comparative analysis - which machine learning algorithm is better in predicting the openness to new opportunities and readiness to career, providing empirical evidence of interrelation between career aspects of students' personality and their impact on readiness to transition.

 The structure of the current paper is as follows. Section 1 is this Introduction. Section 2 describes methods employed in this study, including Balance Wheel, Survey Design, Fuzzy Fuzzy Sets and Buiding of Machine Learning model. Next, Section 3 provides a description of Intelligent System built upon a model and presents the research findings. Finally, Sections 4 and Section 5 present Discussion and
concluding remarks.

\section{Methods} 
The proposed approach for building Intelligent System  is shown in Figure (\ref{fig:methodology}).
\subsection{Balance Wheel}
In the 1960s Paul J. Meyer invented a “Balance Wheel” technique \cite{Swart2022} - a current state analysis tool. The main idea is to choose several primary life areas and assess each one. Visual representation helps better understand which violates “the wheel” and which site needs more attention. The wheel can be divided into any number of segments. However, the best options are 8-10 segments since we are interested in the general view rather than specific details. Each piece has a scale of 10 divisions (1 division - 1 point). A person has to evaluate each area and give an appropriate assessment from  1 (not satisfied) to 10 (completely satisfied). As a result, we obtain life balance visualization. If one of the segments is significantly larger than the rest, it is most developed. And vice versa, drawing down in segments shows you where to focus your attention. The ideal situation is when the wheel is almost smooth; the assessment gap should not exceed 1-2 divisions. 

\begin{figure}[t]
    \centering
    \includegraphics[width=\textwidth]{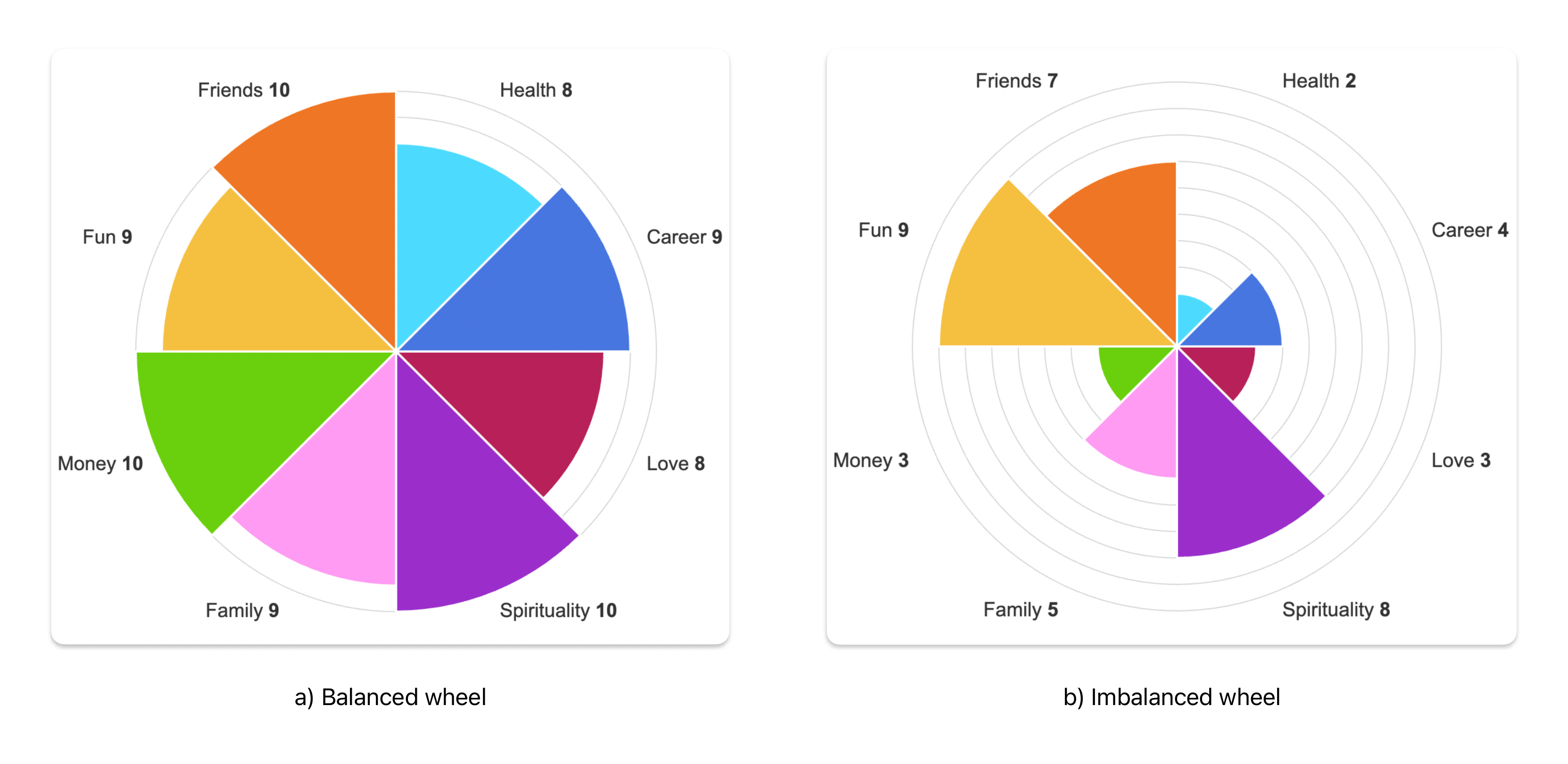}
    \caption{Paul J. Meyer's "Balance Wheel" technique results. }
    \label{fig:wheels}
\end{figure}

Figure \ref{fig:wheels} illustrates different results of the technique. The balanced wheel is what the result looks like for a person with a balanced life. He is in harmony and almost equally effective in different directions. On the other hand, significant discrepancies in assessments indicate that a person is focused on some areas and ignoring others. This imbalance threatens the general dissatisfaction with life.  

\subsection{Survey Design}

A cohort of 47 students, primarily from the Kazakh-British Technical University, participated in the experiment. They sequentially engaged in a survey wherein the questions and their implications were comprehensively explained. The survey was structured based on the concept of the "Balance Wheel" represented in Figure \ref{fig:wheels}. The main aim of the survey was to understand how different parts of a student's life affect the decisions they make.

Constructed via Google Form, the survey encompassed 19 questions, outlined in Table \ref{tab:questionnaire}. Among them, three were personal, namely, the student's name, contact information, and gender. The remaining question spanned topics such as studies, career goals, family relationships, and free time. 
At the end of the survey, participants were asked about their willingness to try new opportunities and their readiness for life changes. Then, we chose the question with opportunities as a target for the experiment.

\begin{figure}[t]
    \centering
    \includegraphics[width=\textwidth]{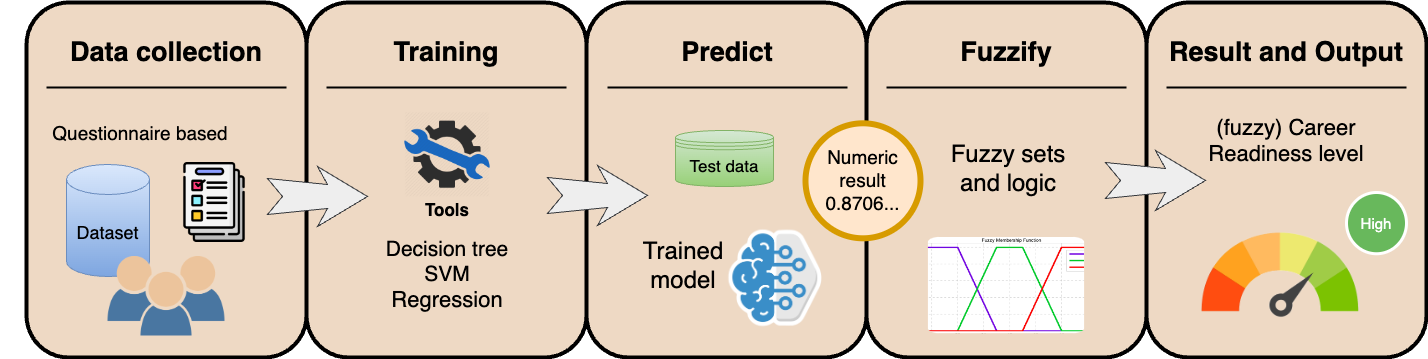}
    \caption{Proposed Methodology}
    \label{fig:methodology}
\end{figure}


\subsection{Survey Analysis}
The analysis of the gathered data was conducted using the Python within the Jupyter Notebook environment. Following data labeling, the primary indicators associated with the data for each specific question were examined. The data exhibited a non-normal distribution, and the presentation of response indicators was skewed to the right, closer to the "10" answer. Evidently, numerous students displayed positive reactions, as illustrated in the \figurename{} \ref{fig:Figure 3}.

\begin{table}[t]
\caption{Survey questionnaire based on the Balance Wheel} 
\label{tab:questionnaire}
\small
\begin{tabular*}{\hsize}{@{\extracolsep{\fill}}lll@{}lll@{}}
\toprule
 & \textbf{Questions of the survey} & \textbf{Label} & \textbf{Balance Wheel Section} \\
1 & Enter your name & Name & General \\
2 & Leave your contacts for feedback ... & PhoneNumber & General \\
3 & Rate your satisfaction with the learning process? & LearningRate & Career \\
4 & Assess the likelihood of finding a job in your specialty after graduation? & WorkingExp & Career \\
5 & Assess the likelihood of receiving an adequate salary after graduation? & SalaryExp & Money \\
6 & Do you have enough free time to spend with family and friends? & FamilyTime & Family \\
7 & Rate how communicative you are? & CommunicationRate & Fun \\
8 & Do you have time for hobbies and hobbies? & HobbyTimeRate & Fun \\
9 & Rate your interest in participating in any business community ... & CommunityRate & Friends \\
10 & Assess your current physical form? & PhysicalFormRate & Health \\
11 & Would you like to improve your physical condition and fitness? & WantToUpPhysicalForm & Health \\
12 & Are you eating healthy enough? & NutritionRate & Health \\
13 & How easy is it for you to find a solution in conflict situations, problems? & ConflictSituations & Love / Career / Friends / Family \\
14 & How much do you think you are in your "comfort zone"? & ComfortZone & Spirituality \\
15 & How open are you to new opportunities and suggestions? & Opportunities & All aspects \\
16 & How much would you like to change something in your life? & ChangeLife & All aspects \\
\end{tabular*}
\end{table}


The Table \ref{table:statistics} provides an overview of the descriptive statistics pertaining to the collected data. Although our dataset encompasses 14 groups (columns), we have centered our analysis on the \textit{'Opportunities'}. It provides insights into students' viewpoints when confronted with novel possibilities. The mean readiness score for students is 8.24, indicating a noteworthy inclination among students to explore new opportunities.

\begin{table}[bt]
\caption{Descriptive Statistics of the collected dataset}
\label{table:statistics}
\begin{tabular*}{\hsize}{@{\extracolsep{\fill}}lll@{}lll@{}lll@{}lll@{}lll@{}lll@{}lll@{}lll@{}}
\toprule
   & \textbf{count} & \textbf{mean} & \textbf{std} & \textbf{min} & \textbf{25\%} & \textbf{50\%} & \textbf{75\%} & \textbf{max} \\
\textbf{Learning Rate}            & 47.0           & 7.13          & 2.18         & 1.0          & 6.0           & 7.0           & 8.5           & 10.0         \\
\textbf{Working Exp}              & 47.0           & 7.59          & 2.04         & 2.0          & 7.0           & 8.0           & 9.0           & 10.0         \\
\textbf{Salary Exp}               & 47.0           & 7.83          & 1.66         & 5.0          & 7.0           & 8.0           & 9.0           & 10.0         \\
\textbf{Family Time}              & 47.0           & 7.11          & 2.18         & 1.0          & 5.5           & 7.0           & 9.0           & 10.0         \\
\textbf{Communication Rate}       & 47.0           & 7.74          & 2.19         & 1.0          & 7.0           & 8.0           & 9.0           & 10.0         \\
\textbf{Hobby Time Rate}          & 47.0           & 6.40          & 2.37         & 1.0          & 5.0           & 7.0           & 8.0           & 10.0         \\
\textbf{Community Rate}           & 47.0           & 6.34          & 2.73         & 1.0          & 4.5           & 7.0           & 8.5           & 10.0         \\
\textbf{Physical Form Rate}       & 47.0           & 6.98          & 2.17         & 1.0          & 6.0           & 7.0           & 8.5           & 10.0         \\
\textbf{Want To Up Physical From} & 47.0           & 9.08          & 1.65         & 1.0          & 8.0           & 10.0          & 10.0          & 10.0         \\
\textbf{Nutrition Rate}           & 47.0           & 5.97          & 2.33         & 1.0          & 5.0           & 6.0           & 7.0           & 10.0         \\
\textbf{Conflict Situations}      & 47.0           & 7.68          & 1.66         & 3.0          & 7.0           & 8.0           & 9.0           & 10.0         \\
\textbf{Comfort Zone}             & 47.0           & 6.51          & 2.26         & 1.0          & 5.5           & 7.0           & 8.0           & 10.0         \\
\textbf{Opportunities}            & 47.0           & 8.17          & 2.04         & 3.0          & 7.0           & 9.0           & 10.0          & 10.0         \\
\textbf{Change Life}              & 47.0           & 8.14          & 1.95         & 1.0          & 7.0           & 8.0           & 10.0          & 10.0         \\
\textbf{GPA}                      & 24.0           & 2.69          & 0.68         & 0.78         & 2.34          & 2.81          & 3.18          & 3.81    \\    
\end{tabular*}
\end{table}

Attributes such as \textit{'LearningRate'} and \textit{'CommunicationRate'} exhibit comparable patterns, with mean scores of 7.22 and 7.84, respectively. The \textit{'SalaryExp'} attribute holds an average score of 7.92. These scores present minor fluctuations from the mean, approximately by a margin of 2 points. This suggests a consistent tendency among students in their responses across distinct question groups. The analysis underscores the students' persistent inclination to embrace new opportunities. The heatmap presented in \figurename{} \ref{fig:correlation_values} illustrates the correlations among the data.


  \begin{figure}[ht]
  \centering
    \begin{minipage}{0.45\linewidth}
        \centering
        \includegraphics[height=2.5in]{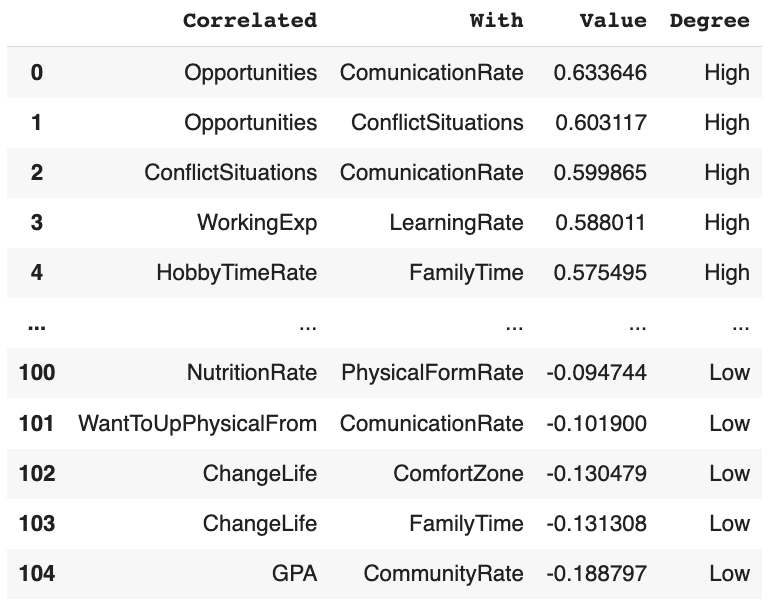}
        \captionof{figure}{Sorted by descending order}
        \label{fig:Figure 5}
    \end{minipage}
    \begin{minipage}{0.45\linewidth}
        \centering
        \includegraphics[height=2.5in]{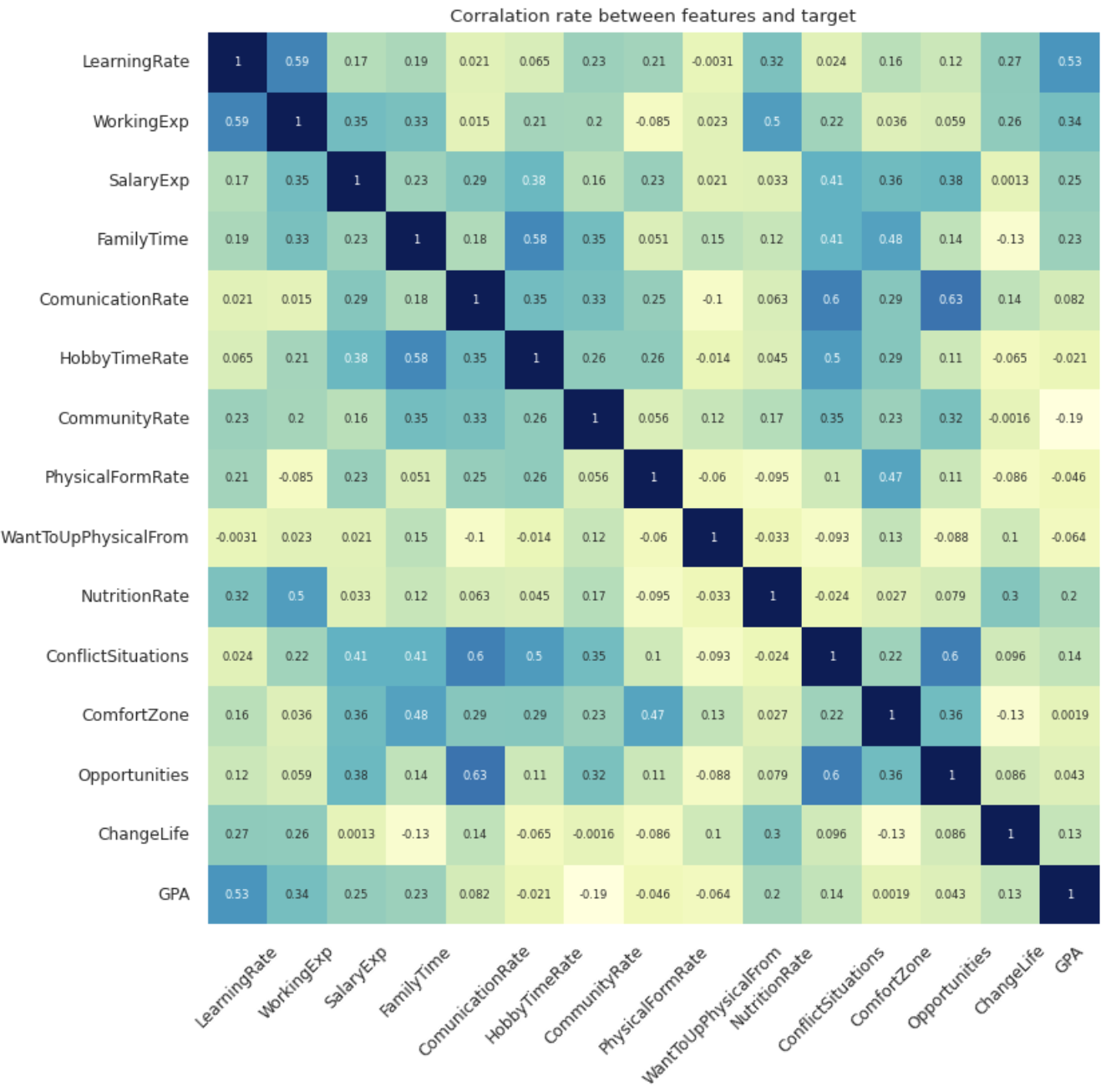}
        \captionof{figure}{Heatmap of correlation between data}
        \label{fig:heatmap}
    \end{minipage}
\end{figure}

Correlation analysis among the gathered data yields interesting insights.  The most significant correlation is 0.65. Specifically, this strong correlation is observed between questions  6 and 15 (see Table \ref{tab:questionnaire}) related to \textit{Sociability }and \textit{Openness} to new opportunities. Next, a correlation of 0.62 exists between \textit{Sociability} (question 6) and \textit{Conflict resolution }(question 13). This correlation indicates that, based on the data, students who possess higher sociability tend to be more adept at resolving conflicts effectively.
    
Subsequently, responses for questions 3 \textit{(LearningRate)} and 4 (\textit{WorkingExp)} exhibit an equal correlation of 0.61 as the responses for questions 6 \textit{(FamilyTime)} and 8 \textit{(HobbyTimeRate)}. This signifies that students who are more engaged in their educational journey are also more confident about their post-graduation job prospects. Additionally, data suggests that if a student prioritizes hobbies, they are more likely to allocate time for personal relationships as well. Responses associated with \textit{Hobby time }and \textit{Conflict resolution}, \textit{Work expectations} and \textit{Healthy diet}, \textit{Sociability} and \textit{Comfort zone} demonstrate correlations of 0.54, 0.53, and 0.52, respectively. this indicates that students who make time for hobbies are more adept at conflict resolution. Furthermore, maintaining a proper diet is linked to heightened confidence in post-university employment. Moreover, those who perceive themselves to be in their "comfort zone" are inclined to dedicate more time to family and friends. The remaining correlations are visible in \figurename{} \ref{fig:Figure 5}.

\subsection{Building a machine learning model}

Based on our established methodology (\ref{fig:methodology}), after data collection we initiated the training of models. Our code implementation commenced after a feature engineering and our proprietary data. Specifically, we analyze data distributions and the correlations between variables and the target feature. This exploration underscored the advantage of selecting features such as \textit{CommunityRate}, \textit{ComfortZone}, \textit{SalaryExp}, and \textit{ComunicationRate} for the depiction of the target variable. This selection was driven by their notable correlation, exceeding 0.32, revealing a substantial relationship between these features and the target variable.

Then, we decided to create a regression model because of the numerical specificity of our dataset. Drawing from the insights presented in this article \cite{statistics_solutions}, our decision hinged on the significance of a correlation values, as in Figure \ref{fig:heatmap}, surpassing 0.3, which indicates a moderate correlation and a discernible connection within this data. As a result, we chose these four features as our variables for regression model. 


  \begin{figure}[t]
  \centering
    \begin{minipage}{0.40\linewidth}
        \centering
        \includegraphics[height=1.75in]{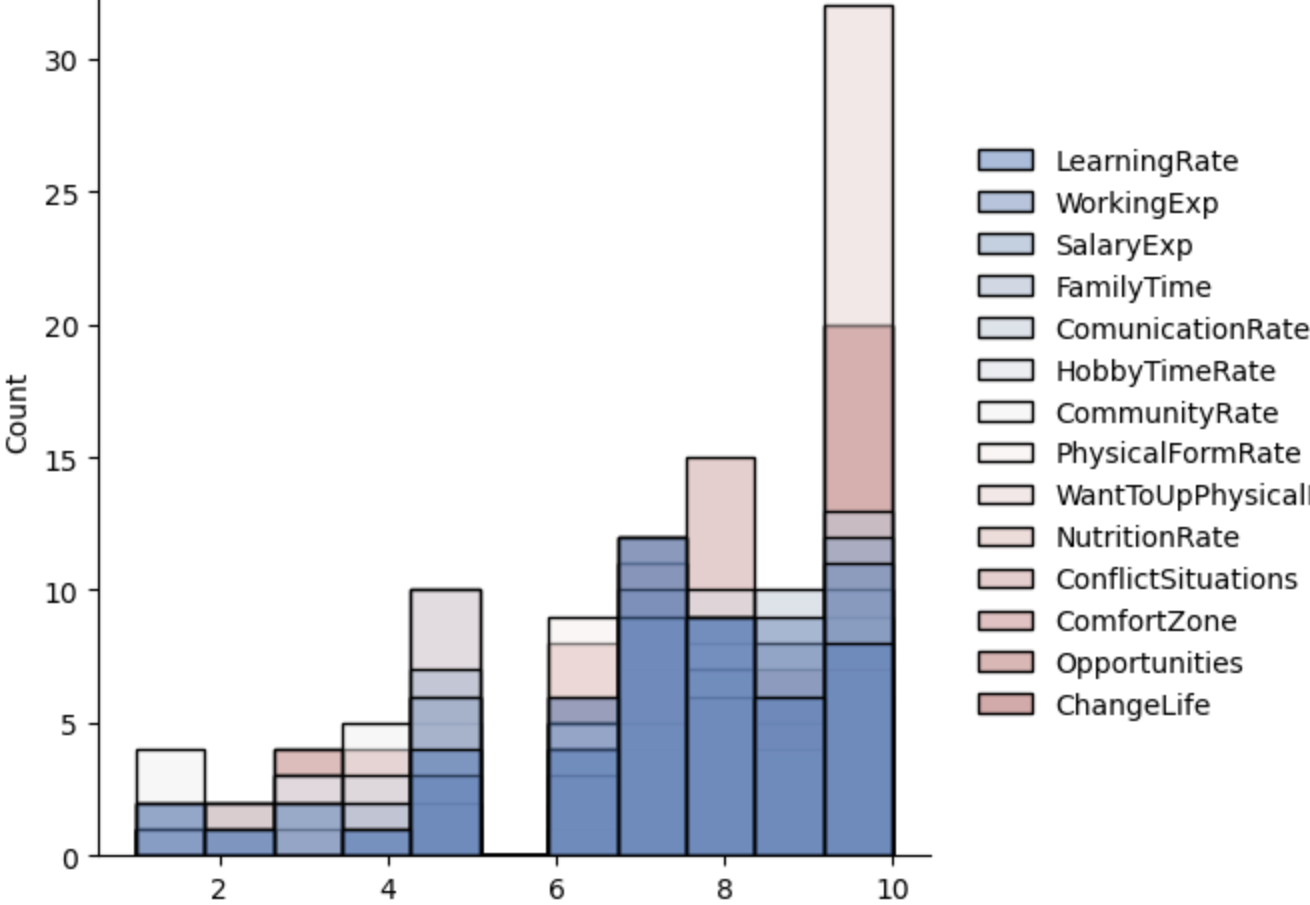}
        \captionof{figure}{Overlay distribution plot of survey responses}
        \label{fig:Figure 3}
    \end{minipage}
    \begin{minipage}{0.55\linewidth}
        \centering
        \includegraphics[height=1.75in]{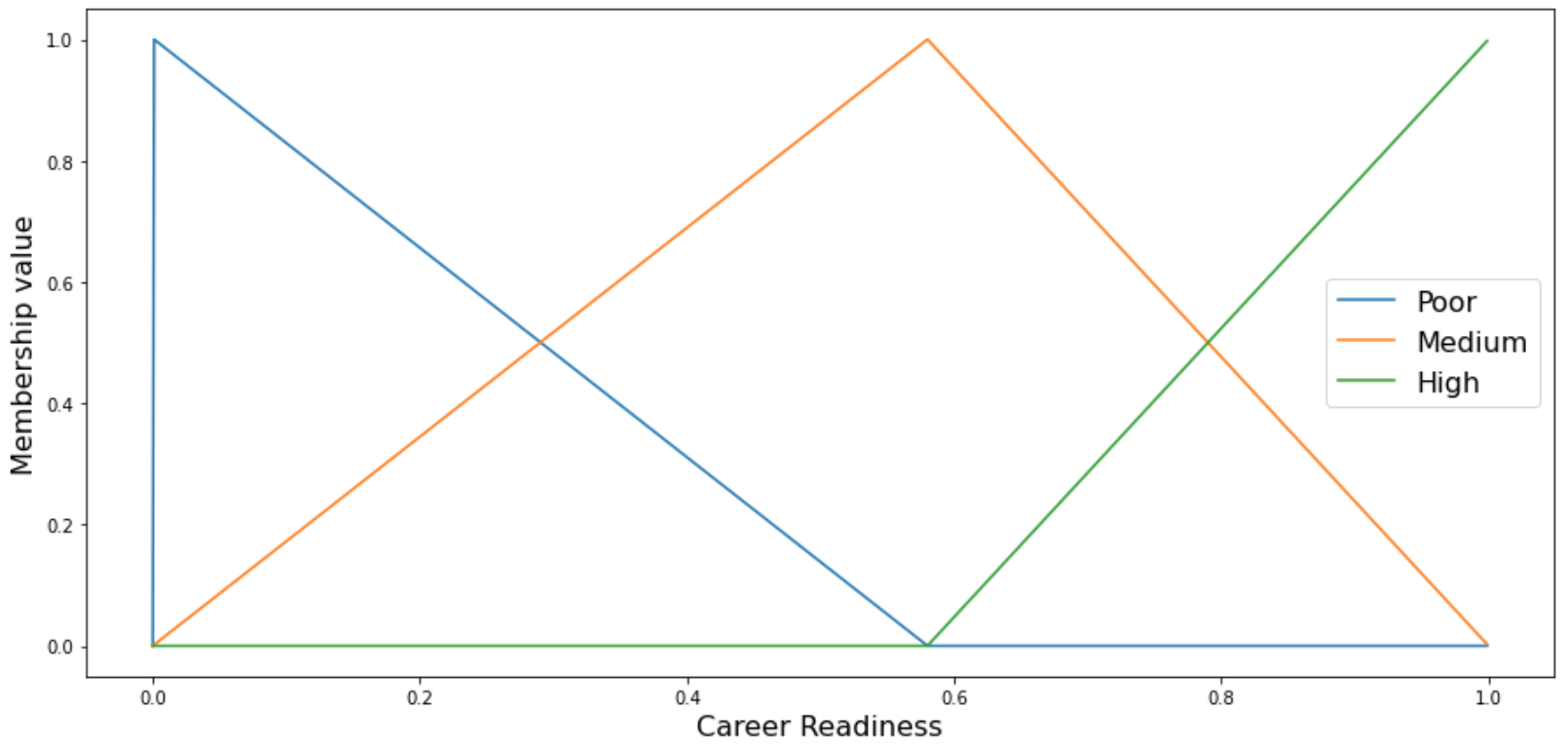}
        \captionof{figure}{Fuzzification of Career Readiness levels}
        \label{fig:fuzzy}
    \end{minipage}
\end{figure}


Next, the data was split into 20\% test  and remaining as a training datasets for the models.
Continuing with that, we build Linear, SVR and Random forest regression models for training with the dataset:
\begin{itemize}
    \item \textit{Linear Regression.} It is a supervised learning algorithm, that is, a model with a single regressor \emph x (independent variables) that has a relationship with a response \emph y (target variable) that is a straight line \cite{linear_book}. The model is defined as $ y =  \beta_0 + \beta_1x + \varepsilon $, where the constant values of intercept $\beta_0$ and scope $\beta_1$ are unknown, and $\varepsilon$ is the random error. Linear regression aims to find the best-fit line that minimizes the sum of squared errors between the predicted and actual values. This line can then be used to make predictions on new data.
    \item \textit{Support Vector Regression (SVR)}. SVR builds on Support Vector Machines (SVM) principles and aims to find a non-linear function that best predicts the target variable. It works by mapping the input features to a high-dimensional feature space using kernel functions and then finding the hyperplane that maximizes the margin between predicted and actual values. SVR can handle non-linear relationships and is less sensitive to outliers than traditional regression models.
    \item \textit{Random Forest Regression.} It is a classifier comprising a set of tree-structured classifiers\textit{ {h(x, $\Theta_k$ ), k = 1,...}}, where \textit{{$\Theta_k$}} denotes independent and identically distributed random vectors \cite{random_forest}. Each tree contributes a unit vote to the most prevalent class for a given input x. This technique enhances prediction accuracy and mitigates overfitting by amalgamating multiple decision trees. 
\end{itemize}

\subsection{Fuzzy sets}
Fuzzy sets can represent ambiguous concepts such as height, age, performance, beauty, smartness, and so on \cite{fuzzycw}. First developed by Zadeh \cite{zadeh}, fuzzy sets allow for degrees of membership, denoted by a number between 0 and 1 (the range [0,1]), in contrast to the pair of numbers 0 and 1. This is known as a \textit{membership value}. Next, the membership function (MF) transforms a crisp value to a membership level in a fuzzy set. It is s denoted as $\mu_A(x)$ and can be used to indicate fuzzy sets. 

According to this logic, the term \textit{poor} is regarded as a linguistic value of the variable \textit{Career Readiness}. As a result, it serves almost the same purpose as a number, but it is linguistic. We represent \textit{Career Readiness} as an ordered list of terms of the fuzzy (linguistic) variable \textit{X = "Career Readiness"} using primary linguistic terms \textit{L = $\{$Poor, Medium, High$\}$}, specifying its level. \figurename{}~\ref{fig:fuzzy} below depicts the fuzzy sets for the \textit{Career Readiness}. We did a few experiments and tweaked the fuzzy sets to create a fuzzy partition depending on subjective impression. Fuzzy partition was made based on a few experiments and adjustments based on personal perception \cite{peerj}. For the sake of simplicity, we use triangular membership functions. $\alpha$-cut is a crisp set that includes members of the given fuzzy subset whose values are not less than $\alpha$ for $0 <\alpha \leq 1 $. The  $\alpha$-cut of $A$ is defined as  $[A]_{\alpha} =  \{ x \in X | A(x) \geqslant \alpha  \}$ \citep{zadeh}.
\section{Results}

After assessing three regression models and conducting a thorough analysis of metrics such as MAE, MSE, and RMSE, we found that among all the models, linear regression exhibited the lowest error, as depicted in Table \ref{table: error_measure}. Consequently, we selected it as our target model. Subsequently, we subjectively defined partitions for the fuzzy sets.

\begin{table}[t]
\caption{Measures of error of models}
\label{table: error_measure}
\begin{tabular*}{\hsize}{@{\extracolsep{\fill}}lll@{}lll@{}}
\toprule
& \textbf{MAE}   & \textbf{MSE}   & \textbf{RMSE}  \\
Linear Regression         & 1.026 & 2.232 & 1.494 \\
Support Vector Regression & 1.317 & 3.148 & 1.774 \\
Random Forest Regression  & 1.266 & 3.137 & 1.771 \\
\end{tabular*}
\end{table}

We then proceeded to transform the actual test values and their corresponding predictions from the regression model into three categories: 'High', 'Medium', and 'Low', using the fuzzy set. The next step involved the calculation of scores, including accuracy, precision, recall, and F1 score, as presented in Table \ref{table: fuzzy_results}. Remarkably, the final model attained an accuracy of 0.8125.

\begin{table}[t]
\caption{Accuracy, Precicion, Recall, F1 score Evaluation Results}
\label{table: fuzzy_results}
\begin{tabular*}{\hsize}{@{\extracolsep{\fill}}lll@{}lll@{}}
\toprule
& \textbf{Accuracy}   & \textbf{Precision}   & \textbf{Recall} & \textbf{F1 Score}  \\
Linear Regression Fuzzy model & 0.8125 & 0.8462 & 0.9167 & 0.88 \\
\end{tabular*}
\end{table}

\section{Discussion}

Let's analyze how our current research outcomes contrast with findings from other investigations. Earlier, there were similar studies, but Bretz et al. primarily focused on the ineffectiveness of the GPA as an indicator of future success. They believe GPA is too subject and situation-specific to be a valid and reliable predictor of job success \cite{Bretz1989}. 


Our results align with later studies that believe a significant positive relationship between higher GPAs and earnings \cite{Afarian2003}.
The students in our survey are more confident that they can find a job with the norms of the GPA. 

Furthermore, our results support the observations from the study regarding learning, and intellectual skills. Lex Borghans et al.\cite{Borghans2016} found that personality is one of the most significant success factors. The findings emphasize that non-cognitive skills can play an essential role in success. 

We delve into a more specific subject, examining not just employment and personality, but also communication, as well as conflict resolution. Post-college employment and income are measurable indicators of higher education success. Particularly among students attending 4-year colleges, a connection exists between GPA and various factors. These factors include the amount of time dedicated to class preparation, arriving to classes with readiness, active participation by asking questions in class, providing tutoring to peers, timely feedback from faculty, cultivating a strong rapport with instructors, and offering a positive assessment of their overall college learning journey. \cite{Kuh2006} This way, students develop their communication skills and conflict resolution abilities.



\section{Conclusion}
The study addresses the critical gap in assessing students' readiness for career and life beyond academia, a dimension that is often overlooked by universities despite its significant implications. This is done by analyzing the influence of factors on university students' change readiness, particularly in terms of career development, and building an ML-based intelligent system using survey-derived datasets. The developed intelligent system offers universities a valuable tool to foster better-equipped graduates who are prepared for the dynamic demands of the modern world. 

Concerning technical analysis, the linear regression algorithm and decision tree demonstrated minimal errors in predicting students' openness to new opportunities and career readiness. The analysis highlighted communication as a crucial factor influencing openness to opportunities and conflict resolution. Allocating time for family and friends not only accommodates hobbies but also provides a sense of comfort. Moreover, evaluating students' post-graduation employability hinges on their learning process and adequate nutrition.

Our approach has certain limitations, primarily due to the relatively small number of participants in our survey. To address this, we intend to involve a larger sample size, thus expanding our training dataset and improving the overall accuracy of our system. 
As for future works, we plan to evaluate the system's effectiveness not only by using the testing dataset but also by applying it to evaluate the readiness of the 2023 graduating class of the KBTU. 





\bibliographystyle{elsarticle-harv}
\bibliography{export}

\begin{thebibliography}{26}
\expandafter\ifx\csname natexlab\endcsname\relax\def\natexlab#1{#1}\fi
\providecommand{\url}[1]{\texttt{#1}}
\providecommand{\href}[2]{#2}
\providecommand{\path}[1]{#1}
\providecommand{\DOIprefix}{doi:}
\providecommand{\ArXivprefix}{arXiv:}
\providecommand{\URLprefix}{URL: }
\providecommand{\Pubmedprefix}{pmid:}
\providecommand{\doi}[1]{\href{http://dx.doi.org/#1}{\path{#1}}}
\providecommand{\Pubmed}[1]{\href{pmid:#1}{\path{#1}}}
\providecommand{\bibinfo}[2]{#2}
\ifx\xfnm\relax \def\xfnm[#1]{\unskip,\space#1}\fi
\bibitem[{Afarian and Kleiner(2003)}]{Afarian2003}
\bibinfo{author}{Afarian, R.}, \bibinfo{author}{Kleiner, B.H.},
  \bibinfo{year}{2003}.
\newblock \bibinfo{title}{The relationship between grades and career success}.
\newblock \bibinfo{journal}{Management Research News} \bibinfo{volume}{26},
  \bibinfo{pages}{42--51}.
\newblock \DOIprefix\doi{10.1108/01409170310783781}.
\bibitem[{Al-Tameemi and Xue(2019)}]{retention}
\bibinfo{author}{Al-Tameemi, G.}, \bibinfo{author}{Xue, J.},
  \bibinfo{year}{2019}.
\newblock \bibinfo{title}{Towards an intelligent system to improve student
  engagement and retention}.
\newblock \bibinfo{journal}{Procedia Computer Science} \bibinfo{volume}{151},
  \bibinfo{pages}{1120--1127}.
\newblock \DOIprefix\doi{10.1016/j.procs.2019.04.159}. \bibinfo{note}{the 10th
  Int-l Conf. on Ambient Systems, Networks and Techn-s/The 2nd Int-l Conf. on
  Emerging Data and Ind. 4.0}.
\bibitem[{Bhatnagar et~al.(2023)Bhatnagar, Agarwal and Sharma}]{mental}
\bibinfo{author}{Bhatnagar, S.}, \bibinfo{author}{Agarwal, J.},
  \bibinfo{author}{Sharma, O.R.}, \bibinfo{year}{2023}.
\newblock \bibinfo{title}{Detection and classification of anxiety in university
  students through the application of machine learning}.
\newblock \bibinfo{journal}{Procedia Computer Science} \bibinfo{volume}{218},
  \bibinfo{pages}{1542--1550}.
\newblock \DOIprefix\doi{https://doi.org/10.1016/j.procs.2023.01.132}.
  \bibinfo{note}{international Conference on Machine Learning and Data
  Engineering}.
\bibitem[{Borghans et~al.(2016)Borghans, Golsteyn, Heckman and
  Humphries}]{Borghans2016}
\bibinfo{author}{Borghans, L.}, \bibinfo{author}{Golsteyn, B.H.},
  \bibinfo{author}{Heckman, J.J.}, \bibinfo{author}{Humphries, J.E.},
  \bibinfo{year}{2016}.
\newblock \bibinfo{title}{What grades and achievement tests measure}.
\newblock \bibinfo{journal}{Proceedings of the National Academy of Sciences of
  the United States of America} \bibinfo{volume}{113},
  \bibinfo{pages}{13354--13359}.
\newblock \DOIprefix\doi{10.1073/pnas.1601135113}.
\bibitem[{Bouslama et~al.(2014)Bouslama, Housley and Steele}]{dubai}
\bibinfo{author}{Bouslama, F.}, \bibinfo{author}{Housley, M.},
  \bibinfo{author}{Steele, A.}, \bibinfo{year}{2014}.
\newblock \bibinfo{title}{A fuzzy logic-based emotional intelligence framework
  for evaluating and orienting new students at hct dubai colleges}, in:
  \bibinfo{booktitle}{2014 14th International Conference on Hybrid Intelligent
  Systems}, pp. \bibinfo{pages}{85--90}.
\newblock \DOIprefix\doi{10.1109/HIS.2014.7086177}.
\bibitem[{Breiman(2001)}]{random_forest}
\bibinfo{author}{Breiman, L.}, \bibinfo{year}{2001}.
\newblock \bibinfo{title}{Random forests}.
\newblock \bibinfo{journal}{Machine Learning} \bibinfo{volume}{45},
  \bibinfo{pages}{5--32}.
\newblock \URLprefix \url{https://doi.org/10.1023/A:1010933404324},
  \DOIprefix\doi{10.1023/A:1010933404324}.
\bibitem[{Bretz()}]{Bretz1989}
\bibinfo{author}{Bretz, R.D.}, .
\newblock \bibinfo{title}{College grade point average as α predictor of adult
  success: A meta-analytic review and some additional evidence} .
\bibitem[{Kaklauskas et~al.(2010)Kaklauskas, Zavadskas, Pruskus, Vlasenko,
  Seniut, Kaklauskas, Matuliauskaite and Gribniak}]{bio}
\bibinfo{author}{Kaklauskas, A.}, \bibinfo{author}{Zavadskas, E.},
  \bibinfo{author}{Pruskus, V.}, \bibinfo{author}{Vlasenko, A.},
  \bibinfo{author}{Seniut, M.}, \bibinfo{author}{Kaklauskas, G.},
  \bibinfo{author}{Matuliauskaite, A.}, \bibinfo{author}{Gribniak, V.},
  \bibinfo{year}{2010}.
\newblock \bibinfo{title}{Biometric and intelligent self-assessment of student
  progress system}.
\newblock \bibinfo{journal}{Computers `I\&' Education} \bibinfo{volume}{55},
  \bibinfo{pages}{821--833}.
\newblock \DOIprefix\doi{10.1016/j.compedu.2010.03.014}.
\bibitem[{Kozakai et~al.(2022)Kozakai, Kobayashi, Wenxuan and
  Watanabe}]{python}
\bibinfo{author}{Kozakai, R.}, \bibinfo{author}{Kobayashi, T.},
  \bibinfo{author}{Wenxuan, Z.}, \bibinfo{author}{Watanabe, Y.},
  \bibinfo{year}{2022}.
\newblock \bibinfo{title}{Tendency analysis of python programming classes for
  junior and senior high school students}.
\newblock \bibinfo{journal}{Procedia Computer Science} \bibinfo{volume}{207},
  \bibinfo{pages}{4603--4612}.
\newblock \DOIprefix\doi{https://doi.org/10.1016/j.procs.2022.09.524}.
  \bibinfo{note}{knowledge-Based and Intelligent Information `I\&' Engineering
  Systems: Proceedings of the 26th International Conference KES2022}.
\bibitem[{Kuh et~al.(2006)Kuh, Kinzie, Buckley, Bridges and Hayek}]{Kuh2006}
\bibinfo{author}{Kuh, G.D.}, \bibinfo{author}{Kinzie, J.},
  \bibinfo{author}{Buckley, J.A.}, \bibinfo{author}{Bridges, B.K.},
  \bibinfo{author}{Hayek, J.C.}, \bibinfo{year}{2006}.
\newblock \bibinfo{title}{What matters to student success: A review of the
  literature commissioned report for the national symposium on postsecondary
  student success: Spearheading a dialog on student success} .
\bibitem[{Kılıç et~al.(2022)Kılıç, Karakuş and Alptekin}]{turks}
\bibinfo{author}{Kılıç, A.C.}, \bibinfo{author}{Karakuş, A.},
  \bibinfo{author}{Alptekin, E.}, \bibinfo{year}{2022}.
\newblock \bibinfo{title}{Prediction of university students’ subjective
  well-being with sleep and physical activity data using classification
  algorithms}.
\newblock \bibinfo{journal}{Procedia Computer Science} \bibinfo{volume}{207},
  \bibinfo{pages}{2648--2657}.
\newblock \DOIprefix\doi{https://doi.org/10.1016/j.procs.2022.09.323}.
  \bibinfo{note}{knowledge-Based and Intelligent Information `I\&' Engineering
  Systems: Proc. of the 26th Int-l Conf. KES2022}.
\bibitem[{Luis et~al.(2017)Luis, Llamas-Nistal and Iglesias}]{new1}
\bibinfo{author}{Luis, R.M.M.F.}, \bibinfo{author}{Llamas-Nistal, M.},
  \bibinfo{author}{Iglesias, M.J.F.}, \bibinfo{year}{2017}.
\newblock \bibinfo{title}{Enhancing learners' experience in e-learning based
  scenarios using intelligent tutoring systems and learning analytics: First
  results from a perception survey}, in: \bibinfo{booktitle}{2017 12th Iberian
  Conference on Information Systems and Technologies (CISTI)}, pp.
  \bibinfo{pages}{1--4}.
\newblock \DOIprefix\doi{10.23919/CISTI.2017.7975976}.
\bibitem[{Montgomery et~al.(2013)Montgomery, Peck and Vining}]{linear_book}
\bibinfo{author}{Montgomery, D.}, \bibinfo{author}{Peck, E.},
  \bibinfo{author}{Vining, G.}, \bibinfo{year}{2013}.
\newblock \bibinfo{title}{Introduction to Linear Regression Analysis}.
\newblock Wiley Series in Probability and Statistics,
  \bibinfo{publisher}{Wiley}.
\bibitem[{Nordin(2011)}]{org}
\bibinfo{author}{Nordin, N.}, \bibinfo{year}{2011}.
\newblock \bibinfo{title}{The influence of emotional intelligence, leadership
  behaviour and organizational commitment on organizational readiness for
  change in higher learning institution}.
\newblock \bibinfo{journal}{Procedia - Social and Behavioral Sciences}
  \bibinfo{volume}{29}, \bibinfo{pages}{129--138}.
\newblock \URLprefix
  \url{https://www.sciencedirect.com/science/article/pii/S1877042811026784},
  \DOIprefix\doi{https://doi.org/10.1016/j.sbspro.2011.11.217}.
  \bibinfo{note}{the 2nd International Conference on Education and Educational
  Psychology 2011}.
\bibitem[{Shamoi et~al.(2022)Shamoi, Turdybay, Shamoi, Akhmetov, Jaxylykova and
  Pak}]{peerj}
\bibinfo{author}{Shamoi, E.}, \bibinfo{author}{Turdybay, A.},
  \bibinfo{author}{Shamoi, P.}, \bibinfo{author}{Akhmetov, I.},
  \bibinfo{author}{Jaxylykova, A.}, \bibinfo{author}{Pak, A.},
  \bibinfo{year}{2022}.
\newblock \bibinfo{title}{Sentiment analysis of vegan related tweets using
  mutual information for feature selection}.
\newblock \bibinfo{journal}{PeerJ Computer Science} \bibinfo{volume}{8},
  \bibinfo{pages}{e1149}.
\newblock \DOIprefix\doi{10.7717/peerj-cs.1149}.
\bibitem[{Shamoi and Inoue(2012)}]{fuzzycw}
\bibinfo{author}{Shamoi, P.}, \bibinfo{author}{Inoue, A.},
  \bibinfo{year}{2012}.
\newblock \bibinfo{title}{Computing with words for direct marketing support
  system}, in: \bibinfo{booktitle}{Midwest Artificial Intelligence and
  Cognitive Science Conference}.
\newblock \URLprefix \url{http://ceur-ws.org/Vol-841/submission_36.pdf}.
\bibitem[{Sheeba and Krishnan(2018)}]{fuzzycool}
\bibinfo{author}{Sheeba, T.}, \bibinfo{author}{Krishnan, R.},
  \bibinfo{year}{2018}.
\newblock \bibinfo{title}{Semantic predictive model of student dynamic profile
  using fuzzy concept}.
\newblock \bibinfo{journal}{Procedia Computer Science} \bibinfo{volume}{132},
  \bibinfo{pages}{1592--1601}.
\newblock \DOIprefix\doi{https://doi.org/10.1016/j.procs.2018.05.124}.
  \bibinfo{note}{int-l Conf. on Computational Intelligence and Data Science}.
\bibitem[{Singh and Singh(2016)}]{case}
\bibinfo{author}{Singh, R.P.}, \bibinfo{author}{Singh, K.},
  \bibinfo{year}{2016}.
\newblock \bibinfo{title}{Design and research of data analysis system for
  student education improvement (case study: Student progression system in
  university)}, in: \bibinfo{booktitle}{2016 International Conference on
  Micro-Electronics and Telecommunication Engineering (ICMETE)}, pp.
  \bibinfo{pages}{508--512}.
\newblock \DOIprefix\doi{10.1109/ICMETE.2016.80}.
\bibitem[{Swart(2022)}]{Swart2022}
\bibinfo{author}{Swart, J.}, \bibinfo{year}{2022}.
\newblock \bibinfo{title}{The wheel of life as a coaching tool to audit life
  priorities} \DOIprefix\doi{10.13140/RG.2.2.25897.67688}.
\bibitem[{Tarik et~al.(2021)Tarik, Aissa and Yousef}]{covid19}
\bibinfo{author}{Tarik, A.}, \bibinfo{author}{Aissa, H.},
  \bibinfo{author}{Yousef, F.}, \bibinfo{year}{2021}.
\newblock \bibinfo{title}{Artificial intelligence and machine learning to
  predict student performance during the covid-19}.
\newblock \bibinfo{journal}{Procedia Computer Science} \bibinfo{volume}{184},
  \bibinfo{pages}{835--840}.
\newblock \DOIprefix\doi{https://doi.org/10.1016/j.procs.2021.03.104}.
  \bibinfo{note}{the 12th Int-l Conf. on Ambient Systems, Networks and
  Technologies (ANT) / The 4th Int-l Conf. on Emerging Data and Industry 4.0
  (EDI40) / Affiliated Workshops}.
\bibitem[{Thai-Nghe et~al.(2010)Thai-Nghe, Drumond, Krohn-Grimberghe and
  Schmidt-Thieme}]{perf}
\bibinfo{author}{Thai-Nghe, N.}, \bibinfo{author}{Drumond, L.},
  \bibinfo{author}{Krohn-Grimberghe, A.}, \bibinfo{author}{Schmidt-Thieme, L.},
  \bibinfo{year}{2010}.
\newblock \bibinfo{title}{Recommender system for predicting student
  performance}.
\newblock \bibinfo{journal}{Procedia Computer Science} \bibinfo{volume}{1},
  \bibinfo{pages}{2811--2819}.
\newblock \DOIprefix\doi{https://doi.org/10.1016/j.procs.2010.08.006}.
  \bibinfo{note}{proc. of the 1st Workshop on Recommender Systems for Techn.
  Enhanced Learning (RecSysTEL 2010)}.
\bibitem[{Tiwari et~al.(2021)Tiwari, Srivastava and Upadhyay}]{ready}
\bibinfo{author}{Tiwari, S.}, \bibinfo{author}{Srivastava, S.K.},
  \bibinfo{author}{Upadhyay, S.}, \bibinfo{year}{2021}.
\newblock \bibinfo{title}{An analysis of students' readiness and facilitators'
  perception towards e-learning using machine learning algorithms}, in:
  \bibinfo{booktitle}{2021 2nd Int-l Conf. on Intel. Eng. and Man-t}, pp.
  \bibinfo{pages}{504--509}.
\newblock \DOIprefix\doi{10.1109/ICIEM51511.2021.9445296}.
\bibitem[{Wu(2017)}]{success}
\bibinfo{author}{Wu, X.}, \bibinfo{year}{2017}.
\newblock \bibinfo{title}{Main factor analysis of influencing factors of
  college students' success rate}, in: \bibinfo{booktitle}{2017 International
  Conference on Robots `I\&' Intelligent System (ICRIS)}, pp.
  \bibinfo{pages}{198--201}.
\newblock \DOIprefix\doi{10.1109/ICRIS.2017.56}.
\bibitem[{www.statisticssolutions.com(2023)}]{statistics_solutions}
\bibinfo{author}{www.statisticssolutions.com}, \bibinfo{year}{2023}.
\newblock \bibinfo{title}{Pearson’s correlation coefficient} .
\bibitem[{Xu et~al.(2002)Xu, Wang and Su}]{profiling}
\bibinfo{author}{Xu, D.}, \bibinfo{author}{Wang, H.}, \bibinfo{author}{Su, K.},
  \bibinfo{year}{2002}.
\newblock \bibinfo{title}{Intelligent student profiling with fuzzy models}, in:
  \bibinfo{booktitle}{Proceedings of the 35th Annual Hawaii International
  Conference on System Sciences}, pp. \bibinfo{pages}{8 pp.--}.
\newblock \DOIprefix\doi{10.1109/HICSS.2002.994005}.
\bibitem[{Zadeh(1965)}]{zadeh}
\bibinfo{author}{Zadeh, L.A.}, \bibinfo{year}{1965}.
\newblock \bibinfo{title}{Fuzzy sets}.
\newblock \bibinfo{journal}{Information and Control} \bibinfo{volume}{8},
  \bibinfo{pages}{338--353}.
\newblock \DOIprefix\doi{10.1016/S0019-9958(65)90241-X}.

\end{thebibliography}

\clearpage


\end{document}